\documentclass[journal]{IEEEtran}

\usepackage[amssymb]{SIunits}
\usepackage{graphicx}
\usepackage{amssymb}
\usepackage{amsthm}
\usepackage{multirow}
\usepackage{enumerate}
\usepackage{enumitem}
\usepackage{wrapfig}
\usepackage{seqsplit}

\usepackage{amsmath}
\usepackage[cmintegrals]{newtxmath}
\usepackage{bm} 
\usepackage{footnote}
\usepackage{threeparttable}
\usepackage{epsfig}
\usepackage{graphicx}
\usepackage{amsmath}
\usepackage{amssymb}
\usepackage{balance}
\usepackage{mathtools}
\usepackage{placeins}
\usepackage{pifont}
\usepackage[misc]{ifsym}
\usepackage[super]{nth}

\usepackage{url}            
\usepackage{booktabs}       
\usepackage{amsfonts}       
\usepackage{nicefrac}       
\usepackage{microtype}      
\usepackage{lipsum}
\usepackage[hidelinks,colorlinks=true,urlcolor=blue,citecolor=blue,allcolors=blue]{hyperref}
\usepackage{graphicx}
\usepackage{color}
\usepackage[dvipsnames]{xcolor}
\usepackage{todonotes}
\usepackage{marginnote}
\presetkeys{todonotes}{size=\tiny}{}
\usepackage{tikz}
\usepackage{array}
\usepackage[table]{xcolor}
\usepackage{makecell}

\newcommand{\cmark}{\ding{51}}
\newcommand{\xmark}{\ding{55}}

\newcommand{\PreserveBackslash}[1]{\let\temp=\\#1\let\\=\temp}
\newcolumntype{C}[1]{>{\PreserveBackslash\centering}p{#1}}
\newcolumntype{R}[1]{>{\PreserveBackslash\raggedleft}p{#1}}
\newcolumntype{L}[1]{>{\PreserveBackslash\raggedright}p{#1}}

\newcommand{\rom}[1]{\uppercase\expandafter{\romannumeral #1\relax}}

\newcommand{\fref}[1]{Fig.~\ref{#1}}
\newcommand{\sref}[1]{Section~\ref{#1}}

\graphicspath{%
    {./figures/}%
}

\ifCLASSOPTIONcompsoc
  \usepackage[caption=false,font=footnotesize,labelfont=sf,textfont=sf]{subfig}
\else
  \usepackage[caption=false,font=footnotesize]{subfig}
\fi

\ifCLASSOPTIONcompsoc
  \usepackage[nocompress]{cite}
\else
  \usepackage{cite}
\fi

\ifCLASSINFOpdf
\else
\fi
\usepackage{algorithm}
\usepackage{algorithmic}
\ifCLASSOPTIONcompsoc
 \usepackage[caption=false,font=normalsize,labelfont=sf,textfont=sf]{subfig}
\else
 \usepackage[caption=false,font=footnotesize]{subfig}
\fi
\usepackage{url}

\begin{document}
%


\title{HiCo-Nav: A Deployable Embodied Vision-Language Navigation System with Hierarchical Cognition and Context-Aware Exploration}

%
%

\author{Kuan~Xu$^{1, \star}$,~
        Ruimeng~Liu$^{1, \star}$,~
        Yizhuo~Yang$^{1}$,~
        Denan~Liang$^{1}$,~
        Tongxing~Jin$^{1}$,\\~
        Shenghai~Yuan$^{1}$,~
        Chen~Wang$^{2}$,~\IEEEmembership{Senior Member,~IEEE,}
        Lihua~Xie$^{1}$,~\IEEEmembership{Fellow,~IEEE}
\thanks{{$^\star$}Equal contribution.}
\thanks{$^{1}$Center for Advanced Robotics Technology Innovation (CARTIN), School of Electrical and Electronic Engineering, Nanyang Technological University, Singapore 639798.}
\thanks{$^{2}$Spatial AI \& Robotics Lab, Department of Computer Science and Engineering, University at Buffalo, Buffalo, NY 14260.}
}

%
%

\markboth{}%
{}
%



\maketitle

\begin{abstract}
Bridging the gap between embodied intelligence and embedded deployment remains a key challenge in intelligent robotic systems, where perception, reasoning, and planning must operate under strict constraints on computation, memory, energy, and real-time execution. In vision-and-language navigation (VLN), existing approaches often face a trade-off between reasoning capability and deployment efficiency on real-world platforms.
In this paper, we present a deployable embodied VLN system that achieves both high efficiency and strong high-level reasoning on real-world robots. The system is decomposed into a fast perception-action layer and a deep reasoning layer running asynchronously at different time scales, with a shared memory layer enabling efficient interaction between them. To support long-horizon reasoning, we incrementally construct a compact memory graph and progressively feed decomposed subgraphs into a vision-language model (VLM). Furthermore, we formulate exploration as a Weighted Traveling Repairman Problem (WTRP) by jointly considering reasoning outcomes and the spatial distribution of candidate regions.
Extensive experiments in simulation and real-world environments demonstrate improved navigation success and efficiency over existing VLN approaches while maintaining real-time performance on resource-constrained hardware. Code and additional real-world experiments are available at \textcolor{blue}{\url{https://github.com/xukuanHIT/HiCo-Nav}}.
\end{abstract}


%

%
\IEEEpeerreviewmaketitle

\section{Introduction}

\IEEEPARstart{N}{avigation} is a fundamental capability for autonomous robotic systems, enabling agents to operate effectively in complex environments \cite{wang2025intelligent, liu2025autonomous}. 
In recent years, navigation that supports natural human-robot interaction has attracted increasing attention, as it enables robots to accomplish more intelligent tasks requiring high-level semantic understanding and reasoning \cite{ye2025rpf, zhu2026emobipednav}. 
In this context, vision-language navigation (VLN) has emerged as a key paradigm that bridges human natural language and robotic visual navigation, allowing embodied agents to plan and execute tasks in unseen environments based on language instructions \cite{liu2025aligning}. 
In particular, zero-shot VLN methods have recently gained increasing attention due to their ability to generalize across complex environments and diverse instructions without additional training, making them highly scalable and suitable for real-world deployment \cite{khan2026comprehensive}.

\begin{figure}[t]
    \centering
    \includegraphics[width= 0.95\linewidth]{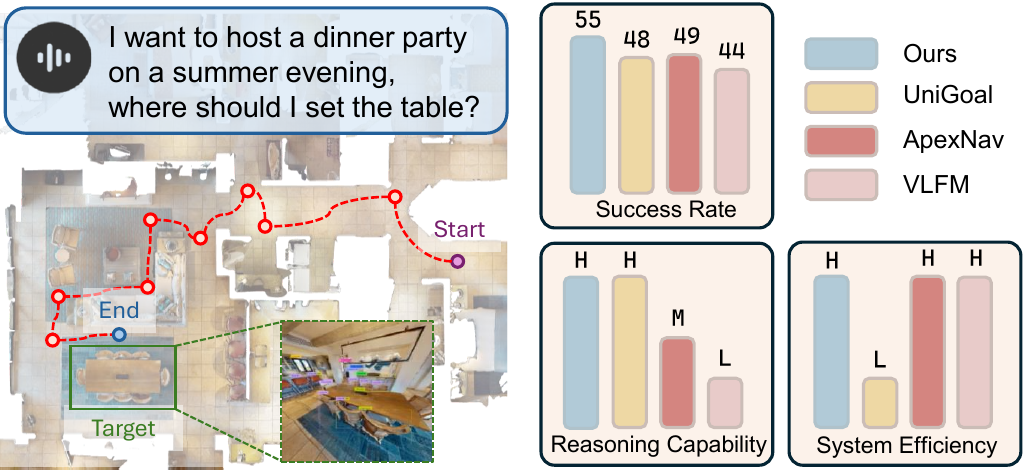}
    \caption{We develop a deployable vision-language navigation system that enables the robot to perform navigation tasks based on language or voice instructions. Both simulation and real-world experiments demonstrate that the proposed system achieves remarkable task success rates, exhibiting strong reasoning capability while maintaining high efficiency.}
    \label{fig:reasoning}
\end{figure}

Despite these advances, existing zero-shot VLN methods still struggle with a fundamental systems-level challenge: navigation intelligence is inherently multi-scale, whereas embedded computation is tightly budgeted. Specifically, in real-world deployment, an embodied agent must react to local geometry and scene environments at high frequency \cite{zhao2024survey}, while also maintaining semantic memory, grounding language to the scene, and making long-horizon decisions under partial observability \cite{liu2025aligning}. 
However, current approaches either prioritize efficiency and deployability with lightweight reasoning modules at the cost of limited scene understanding and reasoning capability \cite{yokoyama2024vlfm, zhang2025apexnav, du2025vl, liu2025handle}, or emphasize strong reasoning through frequent invocation of LLMs/VLMs, which introduces significant latency and undermines real-time responsiveness \cite{yin2024sg, yin2025unigoal, kuang2024openfmnav}.

The most recent works have explored fast-slow paradigms to address this problem \cite{zhou2025fsrvlnfastslowreasoning, wei2025ground}; however, several key challenges remain.
(1) \textit{Tight coupling and blocking}.
In many existing approaches, the fast and slow modules remain tightly coupled: slow reasoning is event-triggered, where the fast module decides when to invoke it and is subsequently blocked by the slow reasoning process \cite{zhou2025fsrvlnfastslowreasoning}. Although more efficient than per-step invocation, this design introduces a trade-off: frequent triggering degrades real-time performance, while conservative triggering may miss targets due to the limited capability of the fast module.
(2) \textit{Insufficient memory utilization}.
The slow module in current methods typically relies on current observations, short-term history, or retrieved candidates, without access to the global, comprehensive memory \cite{zhou2025fsrvlnfastslowreasoning, wei2025ground}. This prevents full utilization of accumulated information and can lead to missed critical cues, especially resulting in false negatives in long-horizon scenarios.
(3) \textit{Greedy exploration redundancy}.
Another challenge is translating fast-slow reasoning into efficient exploration. Specifically, most VLN systems rely on greedy policies that select the most promising region at each step, ignoring the spatial distribution of multiple candidates. This often leads to redundant exploration, oscillatory behavior, and suboptimal trajectories in long-horizon scenarios.

In this work, we propose HiCo-Nav to address these challenges.
Since real-time action and high-level reasoning operate at fundamentally different temporal scales and levels of abstraction, we argue that embodied navigation inherently requires a hierarchical formulation. Accordingly, these components should operate independently and proactively at different layers to enable concurrent acting and reasoning, while an intermediate layer is needed to coordinate them for long-horizon information exchange and propagation without mutual blocking or triggering.
To this end, we decompose the system into a \textit{fast} perception-action layer and a \textit{slow} deep reasoning layer running on separate threads to handle tasks at different time scales, and introduce a memory layer serving as a shared interface. Both the perception-action layer and the reasoning layer proactively interact with and update the shared memory layer.
Meanwhile, for effective utilization of scene memory, both the fast and slow layers should have access to the complete global memory, rather than having the slow module rely on filtered and fragmented memory, as such approximations can lead to semantic omissions and degraded reasoning performance. To this end, we first introduce an integer linear programming (ILP)-based graph pruning scheme to remove redundancy while preserving information completeness. The resulting compact global memory graph is then decomposed into subgraphs and progressively processed by the VLM, enabling efficient and comprehensive long-horizon reasoning.
Furthermore, to address the inefficiency of greedy exploration, we jointly consider reasoning outcomes and the spatial distribution of candidate regions and then formulate exploration as a Weighted Traveling Repairman Problem (WTRP). This results in more consistent exploration and improved long-horizon performance.
In summary, our contributions are as follows:

\begin{itemize}[noitemsep,topsep=0pt]
    \item We present a hierarchical asynchronous architecture for embodied navigation that explicitly bridges high-frequency reactive execution and high-level deep reasoning under constrained onboard computation, enabling both real-time responsiveness and informed decision-making.
    \item We propose a compact global memory graph with progressive subgraph decomposition and an ILP-based pruning scheme, enabling efficient and effective support for both real-time decision-making and deep reasoning.
    \item We develop a context-aware frontier exploration strategy that formulates goal selection as a Weighted Traveling Repairman Problem, enabling globally consistent exploration that minimizes expected time-to-discovery.
    \item We validate the proposed system through extensive simulation and real-world experiments, demonstrating improved navigation efficiency and robustness over existing VLN methods while maintaining real-time performance on resource-constrained embedded platforms.
\end{itemize}

This paper is an extended version of our conference work~\cite{liu2025handle}, with several substantial improvements. First, we introduce a more efficient and flexible memory representation. Unlike~\cite{liu2025handle}, which uses semantic voxels, we incrementally construct a structured and compact cognitive memory graph. This design provides richer semantic information, better supports interaction with VLMs/LLMs, and significantly reduces memory consumption.
Second, we enhance the system’s reasoning capability. While~\cite{liu2025handle} relies on lightweight detectors and CLIP~\cite{radford2021learning}, limiting its ability to handle complex instructions, our hierarchical and decoupled architecture enables the integration of more powerful VLMs for high-level reasoning without blocking real-time execution, thereby significantly improving reasoning capability while preserving real-time performance and deployability.

\section{Related work}\label{sec:related-work}

\subsection{Vision-Language Navigation}

Vision-Language Navigation (VLN) aims to enable agents to interpret natural language instructions and navigate in complex environments. It can be formulated as a language-conditioned sequential decision-making problem under partial observability, requiring joint reasoning over perception, memory, and action~\cite{zhang2024vision}. Early VLN approaches mainly rely on supervised or reinforcement learning, mapping visual observations and instructions to actions via sequence-to-sequence or attention-based models~\cite{hong2021vlnbert}. While effective on benchmarks, these methods often suffer from limited generalization and dependence on large-scale annotated trajectories.
Recent advances improve generalization and reasoning through richer representations and foundation models. Structured representations, such as topological maps, volumetric features, and scene graphs, support long-horizon spatial understanding and planning~\cite{liu2024ver}. Meanwhile, large VLMs and LLMs enhance semantic reasoning and instruction grounding, enabling VLN systems built on pretrained multimodal knowledge~\cite{hong2021vlnbert,zhang2024vision}. Despite these advances, existing methods still struggle to balance high-level reasoning with real-time execution and to achieve efficient exploration under long-horizon and partial observability~\cite{yokoyama2024vlfm,yin2024sg}. These challenges motivate navigation systems that integrate real-time interaction, structured memory, and high-level reasoning for consistent exploration.

\subsection{Goal-Based Exploration}

The development of goal-based exploration has evolved from reactive, end-to-end reinforcement learning to more structured and modular systems that decouple perception, memory, and planning. A representative example is SemExp~\cite{chaplot2020object}, which introduces 2D semantic maps to guide exploration using both geometric and semantic cues. Subsequent works improve generalization with stronger visual representations; for instance, CoW~\cite{gadre2023cow} leverages localized CLIP features for zero-shot object search without environment-specific training.
Recent methods move toward semantically informed exploration, where frontier selection is guided by reasoning rather than purely geometric heuristics. VLFM~\cite{yokoyama2024vlfm} builds probability-driven frontier maps using VLMs, while ApexNav~\cite{zhang2025apexnav} introduces target-centric semantic fusion for adaptive prioritization, highlighting the role of semantic cues in unseen environments.
Beyond 2D representations, richer spatial abstractions and persistent memory have been explored. SG-Nav~\cite{yin2024sg} employs 3D scene graphs with language models for structured reasoning. Generative approaches such as Imagine-before-Go~\cite{zhang2024imagine} infer plausible layouts of unexplored regions, while 3D-Mem~\cite{yang20253d} proposes a unified memory representation for spatial reasoning.
Despite these advances, most methods rely on greedy or locally optimal frontier selection, overlooking the sequential nature of exploration and time-to-discovery, which leads to inefficient long-horizon behavior and motivates globally consistent strategies.

\subsection{LLMs / VLMs for Navigation}

The integration of LLMs/VLMs has significantly enhanced the reasoning capability of embodied agents. Early works such as ZSON~\cite{majumdar2022zson} leverage CLIP embeddings for semantic alignment between goals and observations, followed by extensions like ESC~\cite{zhou2023esc}, which queries LLMs for commonsense priors.
Recent efforts focus on improving reasoning capability and system efficiency. VL-Nav~\cite{du2025vl} and L3MVN~\cite{yu2023l3mvn} propose efficient reasoning frameworks that balance semantic inference with real-time control. To improve generalization across tasks, UniGoal~\cite{yin2025unigoal} unifies diverse navigation goals, including object categories, images, and language instructions, within a single model.
In parallel, several works integrate foundation models into structured decision-making pipelines~\cite{kuang2024openfmnav}. 
GOAT~\cite{chang2023goat} enables universal matching across object categories. WMNav~\cite{nie2025wmnav} incorporates VLM knowledge into world models for predictive planning.
For real-world deployment, Tango~\cite{ziliotto2025tango} advocates training-free agents, and~\cite{cai2024bridging} proposes a pixel-guided skill layer to bridge high-level semantics and low-level control.
Despite these advances, such approaches often face a fundamental trade-off between real-time performance and high-level reasoning capability.
In contrast, our work bridges efficient deployment and high-level reasoning through a hierarchical, asynchronous design. By constructing a compact, queryable memory and formulating exploration as a global optimization problem, it enables efficient, globally consistent navigation while maintaining real-time responsiveness.

\begin{figure*}[t]
    \centering
    \includegraphics[width=0.93\linewidth]{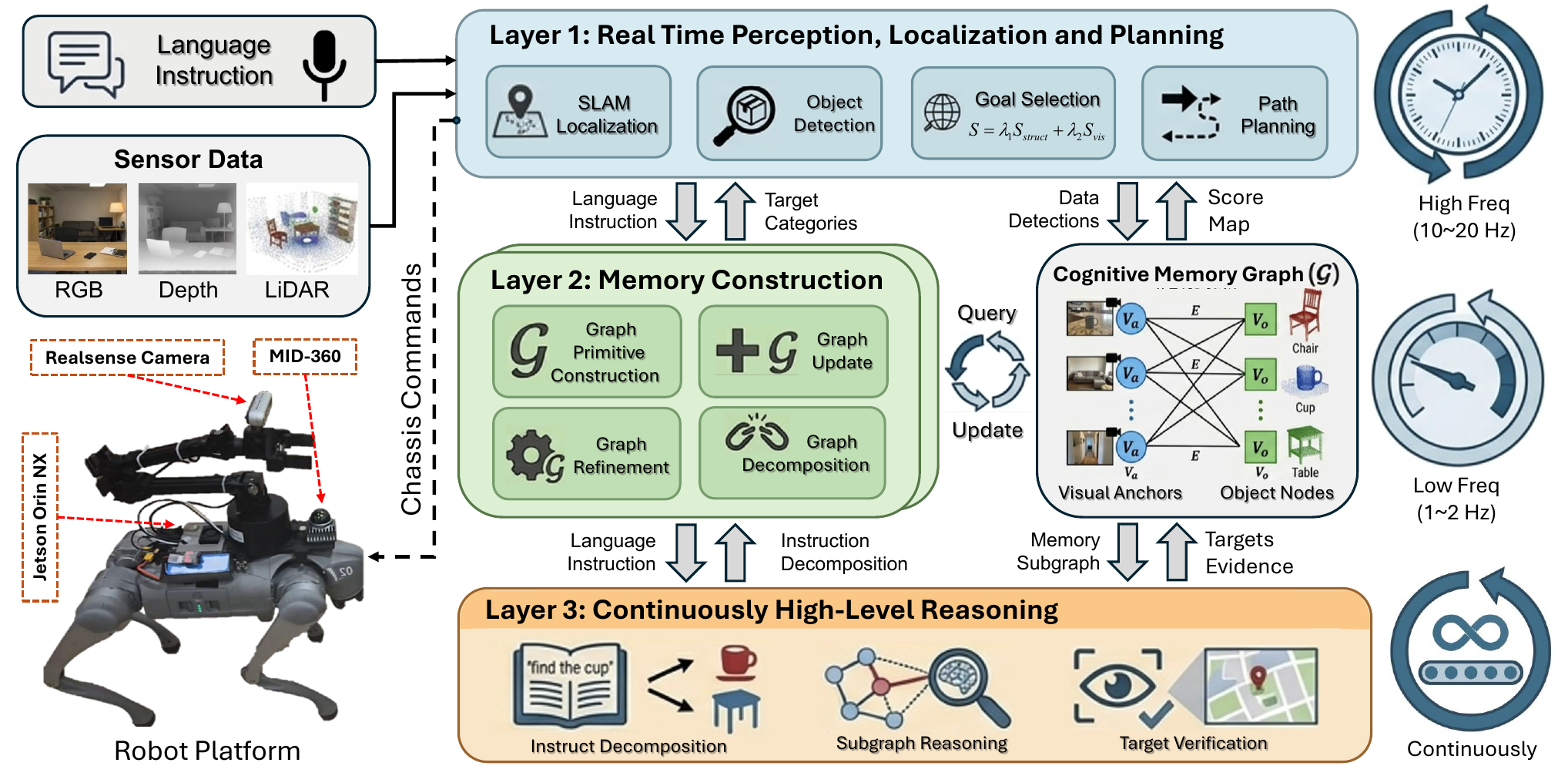}
    \caption{Overview of the proposed system architecture. The framework decouples perception, memory, and reasoning into three layers, enabling real-time sensing, structured memory construction via a cognitive memory graph, and asynchronous VLM-based reasoning for informed navigation.}
    \label{fig:pipeline}
\end{figure*}

\section{Methods}\label{sec:methods}

\subsection{System Framework Design} \label{sec:framework}

The overall system architecture is illustrated in \fref{fig:pipeline}. To address the multi-scale nature of embodied navigation under constrained computation, we design a hierarchical asynchronous architecture that decouples high-frequency perception-action from high-level reasoning while enabling their coordination through a shared memory. Specifically, we decompose the system into three layers.
The first layer is a real-time perception-action layer, including localization, obstacle avoidance, object detection, and both global and local planning, which continuously processes sensory inputs. This layer enables rapid system response and frequent state updates, ensuring timely interaction with dynamic environments.
The second layer is a memory construction layer, where the environment is incrementally abstracted into a cognitive memory graph (CMG). This layer includes scene graph construction, visual anchor optimization, and long-range memory management. It provides an efficient and structured representation of the environment, supporting long-horizon navigation and comprehensive spatial understanding.
The third layer is a high-level reasoning layer. The CMG is decomposed into memory subgraphs, and VLM-based reasoning is continuously performed based on language instructions and memory subgraphs to identify navigation-relevant targets. This layer enables deeper understanding of both instructions and scene context, leading to more informed decision-making.
We next provide a detailed description of each component of the system.

\subsection{Online Memory Construction} \label{sec:memory}

Unlike many existing VLN systems that rely on multi-layer symbolic scene graphs as the primary memory representation, we argue that raw visual observations provide a more compact and information-rich substrate, while being inherently compatible with vision-language models (VLMs). 
To this end, we incrementally construct a cognitive memory graph that jointly models semantic entities and their corresponding visual evidence. The CMG is defined as a bipartite graph
\begin{equation}
\mathcal{G} = (\mathcal{V}_a, \mathcal{V}_o, \mathcal{E}),
\end{equation}
where $\mathcal{V}_a$ denotes the set of visual anchors, $\mathcal{V}_o$ denotes the set of object nodes, and $\mathcal{E}$ represents observation relationships. Each visual anchor corresponds to a selected keyframe that encapsulates a compact visual memory unit, and an edge $(a_i, o_j) \in \mathcal{E}$ indicates that anchor $a_i$ observes object $o_j$.

\paragraph{Graph Primitive Construction}
We employ a MID-360 LiDAR and a RealSense camera for perception. Camera poses are derived from LiDAR-based SLAM \cite{zheng2024fast}, with LiDAR poses transformed into the camera coordinate frame via the calibrated LiDAR-camera extrinsic parameters. For each incoming frame, we perform open-vocabulary object detection using YOLO-World \cite{cheng2024yolo}. A frame is selected as a visual anchor if it satisfies either of the following conditions:
\begin{equation}
\text{(i) } \exists \ \text{new object}, \quad
\text{(ii) } \|\Delta \mathbf{t}\| > \delta_d \ \text{or} \ \Delta \theta > \delta_\theta,
\end{equation}
where $\Delta \mathbf{t}$ and $\Delta \theta$ denote the relative translation and rotation with respect to the last selected anchor.
Selected anchors are forwarded to the memory construction module. For each detected object, we apply Mobile-SAM \cite{mobile_sam} to obtain segmentation masks, and reconstruct its 3D geometry using depth and the camera pose. Each object node $o_j$ is represented by:
\begin{equation}
o_j = \{\mathbf{P}_j, \mathbf{f}_j\},
\end{equation}
where $\mathbf{P}_j$ denotes the 3D point cloud and $\mathbf{f}_j$ is a semantic feature extracted using CLIP \cite{radford2021learning}. 
An edge is established between each object node and the newly selected visual anchor to encode their observation relationship.

\paragraph{Graph Update}
The newly visual anchor node and object nodes are integrated into the global CMG via a similarity-based merging process. Given a new object node $o$ and an existing node $o'$, we compute a joint similarity score:
\begin{equation}
s(o, o') = \lambda_1 \, s_g(o, o') + \lambda_2 \, s_s(o, o'),
\end{equation}
where $s_g$ measures geometric similarity via 3D IoU between point clouds, and $s_s$ denotes semantic similarity computed using CLIP features. The object is merged into the most similar existing node if $s(o,o')$ exceeds a threshold; otherwise, a new node is created. This process incrementally refines the CMG while maintaining a consistent object-level abstraction.

\paragraph{Graph Pruning}
As the CMG grows, the number of visual anchors can become excessive. To maintain a compact yet informative memory, we formulate visual anchor selection as a constrained optimization problem over the CMG.

Let $x_i \in \{0,1\}$ indicate whether visual anchor $a_i \in \mathcal{V}_a$ is selected. The objective is to minimize the total selection cost:
\begin{equation}
\min_{\{x_i\}} \sum_{a_i \in \mathcal{V}_a} c_i x_i,
\end{equation}
subject to object coverage constraints:
\begin{equation}
\sum_{a_i:\,(a_i,o_j)\in\mathcal{E}} x_i \ge r_j, \quad \forall o_j \in \mathcal{V}_o,
\end{equation}
where $c_i$ denotes the cost of anchor $a_i$, and $r_j$ is the required observation count for object $o_j$, defined as
\begin{equation}
r_j = \min \Big(r,\; |\{a_i \mid (a_i,o_j)\in\mathcal{E}\}| \Big).
\end{equation}

This formulation ensures that each object is covered by multiple anchors whenever possible. When $c_i=1$, the objective reduces to minimizing the number of selected anchors; more generally, $c_i$ can encode anchor quality or diversity.
The resulting problem is a weighted set multicover problem, which we solve using integer linear programming (ILP). Compared with heuristic filtering strategies, this formulation explicitly enforces semantic completeness while achieving strong memory compactness. The selected visual anchors thus form a concise yet expressive memory basis, facilitating efficient downstream reasoning and long-horizon navigation.

\subsection{Deep Reasoning} \label{sec:reasoning}

In this section, we present an proactive and asynchronous reasoning module that leverages VLMs over the CMG to achieve efficient yet expressive decision-making under real-time constraints.
Existing methods typically adopt passive, conditionally triggered VLM usage in two forms. The first performs VLM reasoning at every step, achieving strong capability but incurring high latency. The second selectively invokes VLMs via fast-thinking modules, but relies on lightweight models (e.g., CLIP), limiting the understanding of complex instructions and environments and potentially missing navigation targets.

To address these limitations, we propose an active and asynchronous deep reasoning module in this section.
Due to limited computational resources, directly inputting the entire CMG into the VLM is impractical. Therefore, we first partition the cognitive memory map into multiple subgraphs and then employ a separate reasoning thread that incrementally and continuously queries the VLM with each subgraph conditioned on the task instruction. This enables efficient evaluation of candidate regions while maintaining scalability.

\paragraph{Subgraph Decomposition}
Given a CMG $\mathcal{G}=(\mathcal{V}_a,\mathcal{V}_o,\mathcal{E})$, we aim to partition it into a set of subgraphs:
\begin{equation}
\{\mathcal{G}_1, \mathcal{G}_2, \dots, \mathcal{G}_K\},
\end{equation}
where each subgraph $\mathcal{G}_k=(\mathcal{V}_a^k,\mathcal{V}_o^k,\mathcal{E}^k)$ consists of a subset of visual anchors and their associated objects:
\begin{equation}
\mathcal{V}_o^k = \bigcup_{a_i \in \mathcal{V}_a^k} \mathcal{O}(a_i).
\end{equation}

To further improve task efficiency, we perform subgraph decomposition based on the likelihood of containing the target across different regions, enabling the VLM to focus on the most promising areas first.
To this end, we first employ the VLM to decompose the instruction into semantic object cues at the beginning of each task. Specifically, given a task instruction, the VLM extracts two sets of objects: the target object set $\mathcal{O}_t$ and the related object set $\mathcal{O}_r$.
The target objects are those whose observation directly completes the task, while the related objects are strongly semantically associated with the target (e.g., tables and chairs), providing contextual cues that indicate likely locations of the target.
For each target object $o_t \in \mathcal{O}_t$, we construct the subgraph $\mathcal{G}_t = (\mathcal{V}_a^t,\mathcal{V}_o^t,\mathcal{E}^t)$ by including the object itself, the visual anchors that observe it, and all other objects observed by these anchors:
\begin{equation}
\begin{aligned}
\mathcal{V}_a^t &= \{a_i \mid (a_i,o_t)\in\mathcal{E}\}, \\
\mathcal{V}_o^t &= \{\, o_j \mid \exists a_i \in \mathcal{V}_a^t,\ (a_i,o_j)\in\mathcal{E} \,\}, \\
\mathcal{E}^t &= \{\, (a_i,o_j) \mid a_i \in \mathcal{V}_a^t,\ o_j \in \mathcal{V}_o^t \,\} .
\end{aligned}
\end{equation}
The same procedure is applied to construct subgraphs $\mathcal{G}_t$ for all related objects in $\mathcal{O}_r$.
For the remaining visual anchors that are not included in the above subgraphs, each anchor forms an independent subgraph together with the objects it observes.

During the reasoning, we first feed the subgraphs $\mathcal{G}_t$ and $\mathcal{G}_r$ into the VLM. For the remaining subgraphs, we compute a priority score to determine their processing order:
\begin{equation}
S(\mathcal{G}_k) = \sum_{o_i \in \mathcal{V}_o^k} s(o_i, \mathcal{T}),
\end{equation}
where $\mathcal{T}$ denotes the given task instruction, and $s(o_i, \mathcal{T})$ represents the similarity score between object $o_i$ and the task instruction, computed by CLIP.

\begin{figure}[t]
    \centering
    \includegraphics[width= 0.95\linewidth]{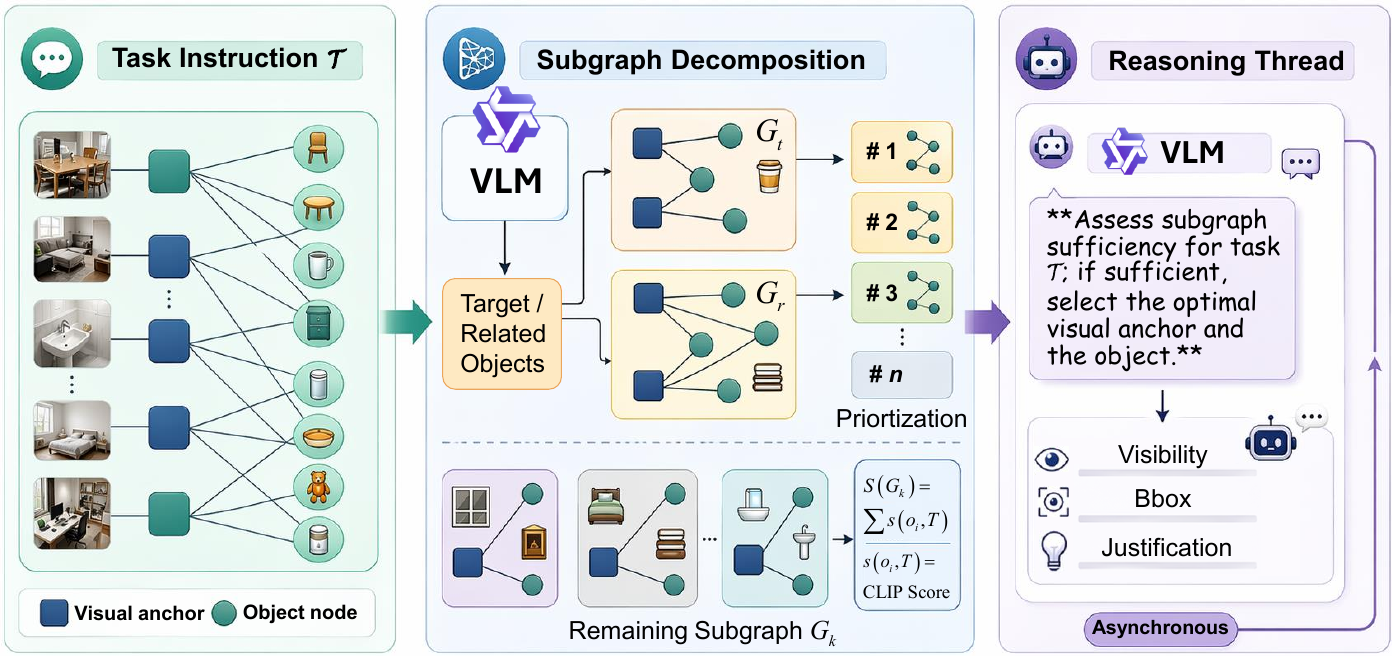}
    \caption{The memory graph is decomposed into subgraphs and prioritized according to their importance. Then the VLM asynchronously processes these subgraphs based on the task instruction for high-level reasoning.}
    \label{fig:reasoning}
\end{figure}

\paragraph{VLM Reasoning}
As shown in \fref{fig:reasoning}, we maintain a separate reasoning thread that sequentially processes subgraphs by querying the Qwen3-Omni API.
We formulate the task-oriented visual reasoning process as a structured prompt-based interaction with the VLM. Given a task instruction $\mathcal{T}$ and subgraphs, the objective is to determine whether the target specified in $\mathcal{T}$ is visible and, if so, to localize it precisely.
Specifically, we guides the VLM to perform three key steps: (1) assess the visibility of the target across all provided visual anchors, (2) select the most relevant visual anchor and localize the target via a bounding box, and (3) provide concise reasoning evidence to support the decision. 
By decoupling VLM reasoning from real-time control and operating over structured subgraphs, the proposed method achieves a balance between efficiency and reasoning capability. The system actively explores the memory space rather than passively reacting to observations, enabling more robust decision-making in complex environments.


\subsection{Context-Aware Frontier Exploration} \label{sec:exploration}

When the target object has not yet been observed, the robot must actively explore unknown regions. We adopt a frontier-based exploration strategy, where each frontier $f_i$ represents a candidate region.
To enable efficient and effective exploration under real-time constraints, we design a \emph{context-aware frontier utility estimation} method based on the cognitive memory graph. The objective is to assign each frontier a utility score $S(f_i)$ that reflects its potential for leading to task completion.

We observe that the utility of a frontier can be inferred from two complementary sources of information:
(i) \emph{scene structure inside the frontier boundary}, and 
(ii) \emph{visual evidence beyond the frontier boundary}.
Most existing approaches typically rely on only one of these sources, leading to incomplete estimation of exploration utility.
To address this limitation, we integrate multiple cues derived from the CMG and visual observations into a unified scoring framework:
\begin{equation}
S(f_i) = \lambda_1 S_{\text{struct}}(f_i) + \lambda_2 S_{\text{vis}}(f_i),
\end{equation}
where $S_{\text{struct}}$ models structural-semantic propagation within the explored region, and $S_{\text{vis}}$ captures visual evidence associated with out-of-boundary observations.

\paragraph{Structural Semantic Propagation}
We model how observed objects provide contextual cues for nearby unexplored regions by constructing a 2D semantic score map from the CMG. The underlying intuition is that object co-occurrence exhibits distance-dependent correlations, i.e., objects are more likely to appear in proximity (e.g., a chair is likely to be near a table), while such contextual influence diminishes with increasing distance.
Specifically, for each object $o_j \in \mathcal{V}_o$, we model its spatial-semantic influence at location $\mathbf{p}$ using a Gaussian-like function centered:
\begin{equation}
\phi_j(\mathbf{p}) = s_j \cdot 
\exp\left(-\frac{\|\mathbf{p} - \mathbf{p}_j\|^2}{2\sigma_j^2}\right),
\end{equation}
where $\mathbf{p}_j$ is the object location, the peak value $s_j$ is defined by the semantic relevance between the object and the task, measured via CLIP similarity, and the variance $\sigma_j$ is determined by the average extent of its 3D bounding box. Formally, each object contributes a spatial field over the map, and the superposition of all such fields forms the semantic score map.


\paragraph{Task-Aware Semantic Modulation}
To incorporate higher-level reasoning into frontier utility estimation, we leverage VLMs to extract task-relevant semantic priors. Compared with lightweight modules (e.g., CLIP or object detectors), VLMs provide stronger understanding of complex tasks and contextual relationships, but incur higher computational cost when invoked during exploration.
To address this trade-off, we adopt a \emph{predictive} rather than \emph{triggered} strategy to incorporate VLM reasoning. Specifically, we leverage task decomposition results obtained from the VLM (\sref{sec:reasoning}), where the VLM is invoked at the beginning of each task to decompose the instruction $\mathcal{T}$ into a set of related objects $\mathcal{O}_r$, serving as semantic cues for guiding exploration.
We use these semantic cues to modulate the structural-semantic score map without additional VLM inference during exploration. For each object in CMG, we apply a semantic gain:
\begin{equation}
\tilde{s}_j =
\begin{cases}
\gamma \cdot s_j, & o_j \in \mathcal{O}_{\text{r}}, \\
s_j, & \text{otherwise},
\end{cases}
\end{equation}
where $\gamma > 1$ is a boosting factor. 

Based on this score map and the spatial location of each frontier $f_i$, we can efficiently evaluate its structural-semantic utility.
The structural score for frontier $f_i$ is computed as:
\begin{equation}
S_{\text{struct}}(f_i) = \sum_{o_j \in \mathcal{V}_o} \tilde{s}_j \cdot \phi_j(\mathbf{p}_i).
\end{equation}

\paragraph{Out-of-Boundary Evidence}
Although frontier regions lie outside the reconstructed map due to limited sensing range, the associated visual observations often retain partial information beyond the boundary. In particular, while depth measurements are truncated, the corresponding 2D images still capture scene content extending into unexplored regions.
To exploit this information, we associate each frontier $f_i$ with a visual anchor $a_i$, which corresponds to the image that observed this boundary during map construction. This provides a visual cue for inferring potential semantic content beyond the explored region.
We define the visual evidence score as the CLIP-based similarity:
\begin{equation}
S_{\text{vis}}(f_i) = s(a_i, \mathcal{T}),
\end{equation}
which measures the semantic alignment between the frontier-associated observation and the task. It captures long-range semantic cues not yet encoded in the CMG, complementing the structural information within the boundary.



\subsection{Goal Selection and Planning} \label{sec:goal_selection}

Given a set of candidate frontiers generated from semantic-aware exploration, the objective of goal selection is to determine an efficient visiting order that minimizes the expected time to discover the task-relevant target.
Most existing approaches adopt greedy strategies that select the frontier with the highest utility at each step.
However, such myopic decisions often lead to inefficient search behaviors, especially in long-horizon tasks. Meanwhile, the agent typically terminates exploration once the target is found, meaning that visiting all candidate viewpoints is unnecessary. Therefore, the goal selection policy should prioritize frontiers with higher likelihood of containing the target while accounting for traversal cost in a global manner.

\paragraph{Problem Formulation}
Let $\mathcal{F} = \{f_1, f_2, \dots, f_n\}$ denote the set of candidate frontiers. Each frontier $f_i$ is associated with a utility score $S(f_i)$ (\sref{sec:exploration}), which reflects its relevance to the task.
We define a visiting sequence $\pi = (\pi_0, \pi_1, \dots, \pi_n)$, where $\pi_0$ denotes the current robot state. 
The cumulative traversal cost up to step $i$ is defined as
\begin{equation}
C_i(\pi) = \sum_{j=1}^{i} M_{\pi_{j-1},\pi_j},
\end{equation}
where $M_{\pi_{j-1},\pi_j}$ is the transition cost defined below.

We formulate the goal selection as minimizing a weighted cumulative latency:
\begin{equation}
\min_{\pi} \sum_{i=1}^{n} W_{\pi_i} \cdot C_i(\pi),
\end{equation}
which corresponds to the Weighted Traveling Repairman Problem (WTRP).
The proposed formulation can be interpreted as minimizing the expected time-to-discovery across candidate frontiers under a weighted distribution over their relative importance. Intuitively, frontiers with larger weights are more likely to contain the targets, and thus should be visited earlier in the sequence. The WTRP objective enforces this prioritization while accounting for traversal cost, leading to a principled trade-off between semantic relevance and motion efficiency.

\paragraph{Utility-Induced Weighting}
To connect semantic utility with global planning, we transform the frontier utility score into the priority weight that reflects its importance in the visiting sequence. Specifically, we first apply min-max normalization to obtain $\bar{S}(f_i) \in [0,1]$, followed by exponential reweighting:
\begin{equation}
W_{\pi_i} = \frac{\exp\big(\beta \cdot \bar{S}(f_i)\big)}{\exp(\beta)},
\end{equation}
where $\beta$ is a temperature-like parameter that controls the contrast of the weighting distribution.
This transformation amplifies differences between candidate frontiers: higher-utility frontiers are assigned disproportionately larger weights, encouraging them to be visited earlier in the sequence, while lower-utility frontiers are suppressed. As $\beta$ increases, the weighting becomes more peaked, emphasizing high-utility candidates and promoting more aggressive, goal-directed behavior. Conversely, smaller values of $\beta$ yield a smoother distribution, allowing more balanced exploration.

\paragraph{Motion Cost Modeling}
To account for both traversal efficiency and robot motion feasibility, we define the pairwise transition cost by considering spatial displacement, heading change, and local frontier structure.
Let $\mathbf{p}_k \in \mathbb{R}^2$ and $\xi_k$ denote the spatial position and heading angle associated with node $k$, respectively. The parameters $\mathbf{v}_{\max}$ and $\dot{\xi}_{\max}$ correspond to the maximum linear and angular velocities of the robot. The nominal travel cost between two nodes $k_r$ and $k_s$ is defined as
\begin{equation}
t(k_r, k_s) = \max \left\{
\frac{d(\mathbf{p}_{k_r}, \mathbf{p}_{k_s})}{\mathbf{v}_{\max}},
\frac{|\xi_{k_r} - \xi_{k_s}|}{\dot{\xi}_{\max}}
\right\},
\end{equation}
where $d(\mathbf{p}_{k_r}, \mathbf{p}_{k_s})$ is the distance between the two nodes. This formulation approximates the time required to complete the transition under bounded translational and rotational motion.

For the transition from the current robot state to a candidate node, we further incorporate two additional terms. The first is a \emph{motion consistency term}, which penalizes candidates that deviate significantly from the current moving direction:
\begin{equation}
c_c(k) = \cos^{-1}
\left(
\frac{(\mathbf{p}_k - \mathbf{p}_0)^\top \mathbf{v}_0}
{\|\mathbf{p}_k - \mathbf{p}_0\| \, \|\mathbf{v}_0\|}
\right),
\end{equation}
where $\mathbf{p}_0$ and $\mathbf{v}_0$ denote the current robot position and velocity, respectively. This term encourages smoother motion and avoids abrupt reorientation at replanning time.
The second is a \emph{local structure term}, introduced for frontier-derived candidates to prioritize small enclosed unexplored regions that may otherwise induce repeated back-and-forth exploration:
\begin{equation}
c_s(k) = \frac{h_k}{h_{\max}},
\end{equation}
where $h_k$ measures the local openness associated with candidate $k$, and $h_{\max}$ is a truncation threshold used for normalization.
To compute $h_k$, we cast a ray from the candidate viewpoint toward the center of its associated frontier cluster. If the ray intersects an observed cell within distance $h_{\max}$, then $h_k$ is defined as the distance from the first hit cell to the cluster center; otherwise, $h_k$ is set to $h_{\max}$. Intuitively, a smaller $h_k$ indicates a more confined unexplored region, which is often beneficial to resolve early before continuing large-scale exploration.

Combining the above terms, the transition cost from the current robot state to node $k \in \{1,2,\dots,n\}$ is defined as
\begin{equation}
M_{0,k} = t(0,k) + w_c \, c_c(k) + w_f \, c_s(k),
\end{equation}
where $w_c$ and $w_f$ are balancing coefficients.
For all other pairs of candidate nodes, we use the nominal travel cost:
\begin{equation}
M_{k_r,k_s} = M_{k_s,k_r} = t(k_r, k_s),
\quad k_r,k_s \in \{1,2,\dots,n\},
\end{equation}
and define $M_{k,0}=0$
since returning to the current state is not considered in the optimization objective.
The resulting optimization problem is solved using a heuristic solver (e.g., LKH), which produces an ordered sequence of candidate frontiers. The first node in the sequence is selected as the mid-term goal and passed to the local planner for execution, while the remaining sequence provides a global guidance for subsequent exploration. 
Once a navigation goal $g$ is selected, the system generates a distance-aware path (using A* for distant targets or direct local planning for nearby ones) and produces smooth, collision-free velocity commands $(\mathbf{v}_t, \dot{\xi}_t)$ via trajectory optimization and dynamic obstacle-aware control.



\section{Experiments} \label{sec:experiments}

\subsection{Experimental Setting} \label{sec:datasets_and_baselines_e}

\textbf{Tasks and Datasets.} We evaluate our method in the Habitat simulator across three navigation tasks and four datasets. For object navigation, we use the HM3D \cite{ramakrishnan2021habitat} and MP3D \cite{Matterport3D} datasets, where language instructions are limited to object categories. The HM3D dataset from the 2022 Habitat Challenge contains 2,000 episodes spanning 20 scenes and 6 goal categories, while the MP3D dataset consists of 2,195 episodes across 11 scenes with 21 goal categories.
To assess performance on open-vocabulary navigation, we use the val unseen split of the HM3D-OVON \cite{yokoyama2024hm3d} dataset, which contains 3,000 episodes across 36 scenes and 49 goal categories.
We further evaluate our method on the TextNav dataset under complex language instructions. Unlike previous datasets, TextNav \cite{sun2024prioritized} specifies a particular object instance and provides a detailed natural language description (e.g., “The bed is located in the corner of the room, with wooden floors and walls surrounding it. There are no other furniture or decorations around the bed.”). This setting requires stronger reasoning capabilities.

\begin{table}[t]
\centering
\caption{Comparison on the object navigation task on the MP3D and HM3D datasets.}
\label{tab:obj_nav_comparison}
\setlength{\tabcolsep}{4.2pt}
\renewcommand{\arraystretch}{1.12}
\begin{tabular}{C{0.24\linewidth}C{0.11\linewidth}|C{0.11\linewidth}C{0.12\linewidth}|C{0.11\linewidth}C{0.11\linewidth}}
\toprule
\multirow{2}{*}{\textbf{Method}}  & \textbf{Training} 
& \multicolumn{2}{c|}{\textbf{MP3D}} 
& \multicolumn{2}{c}{\textbf{HM3D}} \\
& \textbf{Free} 
& \textbf{SR(\%)$\uparrow$} & \textbf{SPL(\%)$\uparrow$} 
& \textbf{SR(\%)$\uparrow$} & \textbf{SPL(\%)$\uparrow$} \\
\midrule
ZSON~\cite{majumdar2022zson}          & \xmark & 15.3 & 4.8 & 25.5 & 12.6 \\
PSL~\cite{sun2024prioritized}         & \xmark & 18.9 & 6.4 & 42.4 & 19.2 \\
Habitat-Web~\cite{ramrakhya2022habitat} & \xmark & 31.6 & 8.5 & 41.5 & 16.0 \\
SGM~\cite{zhang2024imagine}           & \xmark & 37.7 & 14.7 & 60.2 & 30.8 \\
\midrule
CoW~\cite{gadre2023cow}               & \cmark & 9.2  & 4.9 & -    & -   \\
ESC~\cite{zhou2023esc}                & \cmark & 28.7 & 14.2 & 39.2 & 22.3 \\
L3MVN~\cite{yu2023l3mvn}              & \cmark & -    & -   & 50.4 & 23.1 \\
OpenFMNav~\cite{kuang2024openfmnav}   & \cmark & -    & -   & 54.9 & 24.4 \\
TopV-Nav~\cite{zhong2024topv}         & \cmark & 31.9 & 16.1 & 45.9 & 28.0 \\
VLFM~\cite{yokoyama2024vlfm}          & \cmark & 36.4 & 17.5 & 52.5 & 30.4 \\
ApexNav~\cite{zhang2025apexnav}       & \cmark & 39.2 & \cellcolor{green!20} 17.8 & \cellcolor{green!20} 59.6 & \cellcolor{green!50} \textbf{33.0} \\
SG-Nav~\cite{yin2024sg}               & \cmark & 40.2 & 16.0 & 54.0 & 24.9 \\
UniGoal~\cite{yin2025unigoal}         & \cmark & 41.0 & 16.4 & 54.5 & 25.1 \\
WMNav~\cite{nie2025wmnav}             & \cmark & \cellcolor{green!20} 45.4 & 17.2 & 58.1 & 31.2 \\
\textbf{HiCo-Nav (Ours)}              & \cmark & \cellcolor{green!50} \textbf{48.5} & \cellcolor{green!50} \textbf{21.5} & \cellcolor{green!50} \textbf{61.0} & \cellcolor{green!20} 31.8 \\
\bottomrule     
\end{tabular}
\end{table}


\textbf{Evaluation Metrics.}
Following standard practice, we evaluate navigation performance using Success Rate (SR) and Success weighted by Path Length (SPL). The SR is defined as
\begin{equation}
\text{SR} = \frac{N_{\text{success}}}{N_{\text{total}}}, \quad \text{SPL} = \frac{1}{N_{\text{total}}} \sum_{i=1}^{N_{\text{total}}} S_i \cdot \frac{l_i^s}{\max(l_i^s, l_i^a)},
\end{equation}
where $N_{\text{success}}$ and $N_{\text{total}}$ denote the number of successful episodes and the total number of episodes, respectively. $S_i \in \{0,1\}$ indicates whether the $i$-th episode is successful, $l_i^s$ is the shortest path distance to the goal, and $l_i^a$ is the actual path length traversed by the agent.


\subsection{Comparison with State-of-the-art} \label{sec:comparison}

\textbf{Object Navigation.} Table~\ref{tab:obj_nav_comparison} compares our method with prior approaches on the HM3D \cite{ramakrishnan2021habitat} and MP3D \cite{Matterport3D} datasets. HiCo-Nav achieves the best overall performance across both datasets, demonstrating strong navigation success and path efficiency.
On MP3D, HiCo-Nav achieves an SR of 48.5\% and an SPL of 21.5\%, outperforming the previous best method WMNav \cite{nie2025wmnav} by +3.1\% and +4.3\%, respectively. This indicates that our method not only improves task success but also generates more efficient navigation trajectories. On the HM3D dataset, HiCo-Nav achieves the highest SR of 61.0\% while maintaining competitive efficiency (31.8\% SPL), slightly below ApexNav \cite{zhang2025apexnav} (33.0\%) but with a higher success rate (+1.4\%). 
The consistent gains across both datasets indicate that our approach improves both exploration quality and decision-making reliability in standard object navigation scenarios.

\textbf{Open-Vocabulary Navigation.} Table~\ref{tab:open_nav} reports results on the HM3D-OVON \cite{yokoyama2024hm3d} dataset. HiCo-Nav significantly outperforms all baselines, achieving an SR of 52.4\% and an SPL of 20.7\%.
Compared to the strongest training-free baseline TANGO \cite{ziliotto2025tango}, HiCo-Nav improves SR by +16.9\% and SPL by +1.2\%, and exceeds Modular GOAT \cite{khanna2024goat} by +27.5\% in SR. It also surpasses training-based methods such as MTU3D \cite{zhu2025move} by +11.6\% in SR and 8.6\% in SPL. 
These results demonstrate strong generalization capability of our system in open-vocabulary settings. By leveraging structured memory and global exploration, HiCo-Nav can better associate semantic cues with spatial context, leading to more effective search behavior in open-vocabulary navigation tasks.

\begin{table}[t]
\centering
\caption{Comparison on the open-vocabulary navigation task on the HM3D-OVON dataset.}
\label{tab:open_nav}
\setlength{\tabcolsep}{4.2pt}
\renewcommand{\arraystretch}{1.12}
\begin{tabular}{C{0.27\linewidth}C{0.15\linewidth}|C{0.15\linewidth}C{0.15\linewidth}}
\toprule
\multirow{2}{*}{\textbf{Model}}  & \textbf{Training} 
& \multicolumn{2}{c}{\textbf{VAL UNSEEN}} \\
& \textbf{Free} 
& \textbf{SR(\%)$\uparrow$} & \textbf{SPL(\%)$\uparrow$} \\
\midrule
DAGL+OD \cite{yokoyama2024hm3d}       & \xmark &  37.1 & 19.9 \\
Uni-NaVid \cite{zhang2025uninavid}    & \xmark &  39.5 & 19.8 \\
MTU3D \cite{zhu2025move}              & \xmark &  40.8 & 12.1 \\
\midrule
Modular GOAT \cite{khanna2024goat}    & \cmark &  24.9 & 17.2 \\
VLFM \cite{yokoyama2024vlfm}          & \cmark &  35.2 & 19.6 \\
TANGO \cite{ziliotto2025tango}        & \cmark &  35.5 & 19.5 \\
\textbf{HiCo-Nav (Ours)}              & \cmark &  \textbf{52.4} &  \textbf{20.7} \\
\bottomrule     
\end{tabular}
\end{table}

\begin{table}[t]
\centering
\caption{Comparison on the text navigation task on the TextNav dataset.}
\label{tab:text_nav}
\setlength{\tabcolsep}{4.2pt}
\renewcommand{\arraystretch}{1.12}
\begin{tabular}{C{0.27\linewidth}C{0.15\linewidth}|C{0.15\linewidth}C{0.15\linewidth}}
\toprule
\multirow{2}{*}{\textbf{Model}}  & \textbf{Training} 
& \multicolumn{2}{c}{\textbf{TextNav}} \\
& \textbf{Free} 
& \textbf{SR(\%)$\uparrow$} & \textbf{SPL(\%)$\uparrow$} \\
\midrule

ZSON~\cite{majumdar2022zson}           & \xmark &  10.6 & 4.9 \\
PSL \cite{sun2024prioritized}          & \xmark &  16.5 & 7.5 \\
GOAT \cite{chang2023goat}              & \xmark &  17.0 & 8.8 \\
\midrule
CoW~\cite{gadre2023cow}               & \cmark & 1.8  & 1.1   \\
ESC~\cite{zhou2023esc}                & \cmark & 6.5 & 3.7 \\
UniGoal \cite{yin2025unigoal}         & \cmark &  20.2 & 11.4 \\
\textbf{HiCo-Nav (Ours)}              & \cmark &  \textbf{27.8} &  \textbf{12.9} \\
\bottomrule     
\end{tabular}
\end{table}

\textbf{Text Instance Navigation.} Table~\ref{tab:text_nav} presents results on the TextNav \cite{sun2024prioritized} dataset. HiCo-Nav achieves the best performance, with an SR of 27.8\% and an SPL of 12.9\%, outperforming the best training-free baseline UniGoal by +7.6\% and +1.5\%, respectively.
Compared to training-based methods such as PSL \cite{sun2024prioritized} and GOAT \cite{chang2023goat}, the performance gap is more significant, reflecting the increased difficulty of the TextNav task. Unlike category-based navigation, TextNav requires grounding detailed language descriptions to specific object instances within complex scenes.
The consistent improvement indicates that HiCo-Nav is more effective at interpreting fine-grained language instructions and reasoning over scene context. This suggests that the proposed framework can better handle long-horizon decision-making under complex instructions.

\begin{figure}[t]
    \centering
    \includegraphics[width=0.95\linewidth]{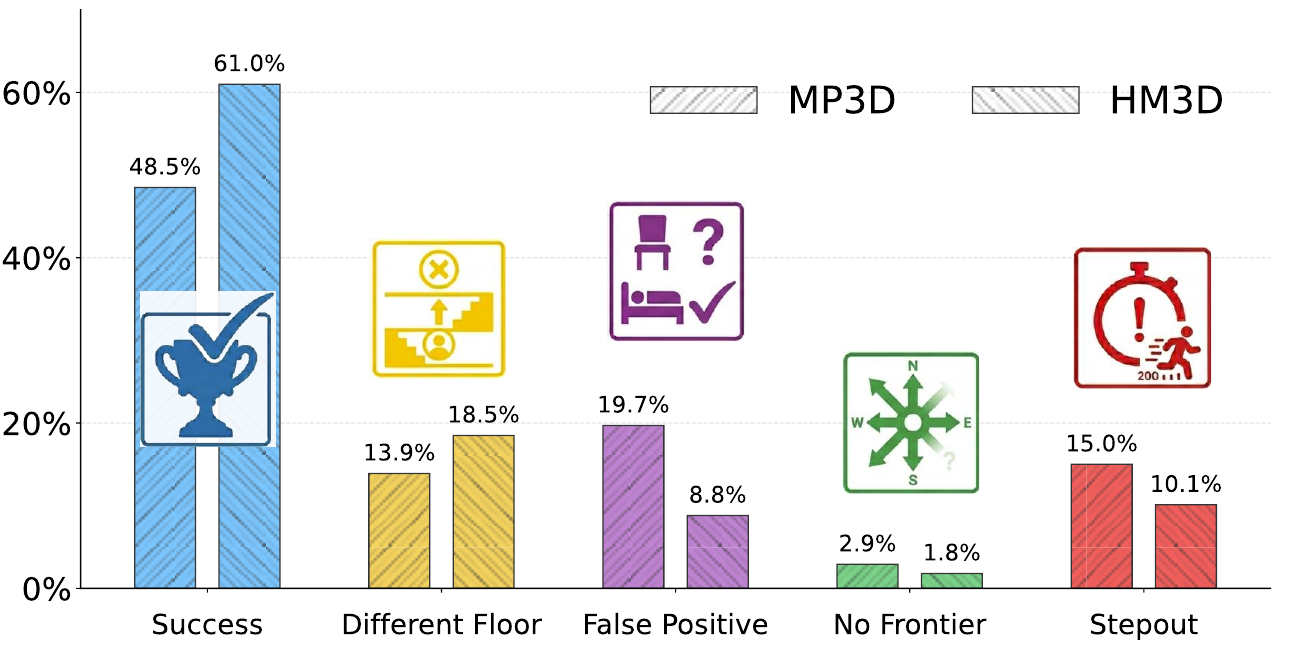}
    \caption{Failure case analysis of our system on the MP3D and HM3D datasets.}
    \label{fig:failure_case}
\end{figure}

\begin{figure*}[t]
    \centering
    \includegraphics[width=0.98\linewidth]{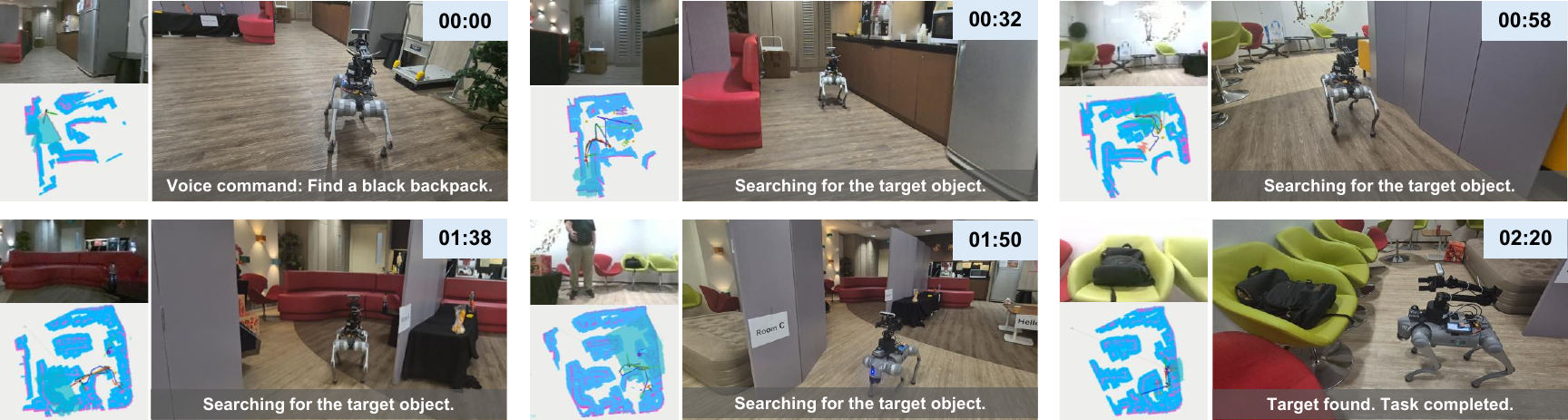}
    \caption{We deploy our system on a quadruped robot, which performs high-level reasoning and real-time navigation based on voice commands.}
    \label{fig:real_world}
    \vspace{-1.0em}
\end{figure*}

\subsection{Failure Case Analysis} \label{sec:failure}

To better understand the limitations of our method, we analyze the outcomes on MP3D and HM3D and categorize them into five types: \textit{Success}, \textit{Different Floor}, \textit{False Positive}, \textit{No Frontier}, and \textit{Stepout}. Specifically, \textit{Success} indicates that the task is successfully completed; \textit{Different Floor} means the agent and target are on different floors; \textit{False Positive} denotes that the agent stops near an incorrect object; \textit{No Frontier} indicates that no unexplored frontiers remain; and \textit{Stepout} means the agent exceeds the step limit (200 steps) without success.

The results are summarized in \fref{fig:failure_case}. We observe that a significant portion of failures is caused by the agent and the target being located on different floors (13.9\% on MP3D and 18.5\% on HM3D). Since our system relies on a 2D exploration map, it is unable to handle cross-floor navigation, which remains an important direction for future improvement.
Another major source of failure is false positives (19.7\% on MP3D and 8.8\% on HM3D). These cases are mainly attributed to two factors. First, annotation errors in the datasets: we observe that some objects belonging to the target category are not labeled as valid targets. Second, perception errors under challenging viewpoints, where visually similar objects may be misclassified (e.g., a sofa misidentified as a bed under extreme viewing angles).
In addition, 15.0\% of failures on MP3D and 10.1\% on HM3D are caused by exceeding the maximum step limit. This typically occurs in large environments where the predefined limit of 200 steps is insufficient for complete exploration.
Finally, a small portion of failures (2.9\% on MP3D and 1.8\% on HM3D) occurs when the agent exhausts all frontiers without locating the target. These cases are mainly due to the target not being observed or failing to be detected even when observed.

\subsection{Efficiency Analysis} \label{sec:efficiency}

Efficiency is essential for robotic applications; therefore, we further evaluate the efficiency of the proposed system. We compare our method with four training-free baselines, including VLFM~\cite{yokoyama2024vlfm}, ApexNav~\cite{zhang2025apexnav}, SG-Nav~\cite{yin2024sg}, and UniGoal~\cite{yin2025unigoal}. Experiments are conducted on the first 100 episodes of HM3D. 
We evaluate three metrics: (1) \textit{task completion time}, (2) \textit{navigation speed} (defined as step distance divided by per-step execution time), and (3) \textit{per-frame processing time}. All experiments are conducted on a platform equipped with an NVIDIA GeForce RTX 5090 and an AMD Ryzen 9 9950X3D.

The results are summarized in Table~\ref{tab:obj_nav_comparison}. In terms of task completion time, our online variant achieves the best performance, completing each task in 21.5\,s on average, which is approximately 33\% faster than VLFM. In addition, our method achieves competitive navigation speed and per-frame processing time, on par with lightweight approaches such as VLFM and ApexNav, while substantially surpassing SG-Nav and UniGoal in overall efficiency.
For the reasoning capability, VLFM relies on lightweight CLIP-based text-image retrieval and thus exhibits limited reasoning ability. ApexNav follows a similar design during navigation, with additional offline LLM-based instruction processing. In contrast, SG-Nav and UniGoal invoke LLMs or VLMs at every step, achieving stronger reasoning at the cost of substantial computational overhead.
Our method adopts an asynchronous VLM-based reasoning strategy, enabling continuous high-level reasoning without blocking real-time execution. As a result, it achieves a favorable balance between efficiency and reasoning capability.

\begin{table}[t]
\centering
\caption{Efficiency analysis.}
\label{tab:obj_nav_comparison}
\setlength{\tabcolsep}{4.2pt}
\renewcommand{\arraystretch}{1.12}
\begin{threeparttable}
\begin{tabular}{C{0.25\linewidth}|C{0.12\linewidth}C{0.15\linewidth}C{0.12\linewidth}|C{0.15\linewidth}}
\toprule
\multirow{2}{*}{\textbf{Method}} & \textbf{Task} & \textbf{Step} & \textbf{Frame}  & \textbf{Reasoning}   \\
& \textbf{Time} $\downarrow$ & \textbf{Speed} $\uparrow$ & \textbf{Time} $\downarrow$  & \textbf{Capability}   \\
\midrule
VLFM~\cite{yokoyama2024vlfm}          & 33.9\second  & 1.00 \meter/\second & 0.25\second  & Low  \\
ApexNav~\cite{zhang2025apexnav}       & 44.4\second  & \textbf{1.67} \meter/\second & \textbf{0.15}\second  & Medium \\
SG-Nav~\cite{yin2024sg}               & 569.3\second & 0.11 \meter/\second & 2.36\second   & High   \\
UniGoal~\cite{yin2025unigoal}         & 298.3\second & 0.23 \meter/\second & 1.07\second   & High   \\
\midrule
\textbf{Ours (offline)\tnote{\dag}}               & 27.8\second & 1.47 \meter/\second & 0.23\second   & High  \\
\textbf{Ours (online)\tnote{\dag}}                & \textbf{21.5}\second & 1.54 \meter/\second & 0.22\second & High  \\
\bottomrule     
\end{tabular}
\begin{tablenotes}
    \footnotesize
    \item[\dag] We evaluate the efficiency of our system using online API calls to Qwen3-Omni and offline locally deployed Qwen3-VL-8B.
\end{tablenotes}
\end{threeparttable}
\end{table}

\subsection{Real-world Experiment} \label{sec:real_world}

We deploy our system on a Unitree quadruped robot equipped with a Mid-360 LiDAR, a RealSense D455 camera, and a Jetson Orin NX. High-level reasoning is performed via online queries to the Qwen3-Omni API.
We use voice commands to instruct the robot to search for target objects. The objects are divided into two groups: large objects (e.g., fire extinguishers, backpacks, and potted plants) and small objects (e.g., bread, carrots, and paper cups). For each group, 20 language instructions are designed.
The results show that our system achieves a success rate of 95\% for large objects and 65\% for small objects. The lower performance on small objects is mainly due to motion-induced vibration of the robot, which causes image blur and reduces detection reliability.
\fref{fig:real_world} shows an example of successful task execution. 

We further evaluate efficiency in real-world deployment. Unlike the simulator setting in Sec.~\ref{sec:efficiency}, observations on the real robot are continuous rather than discretely sampled. To ensure real-time performance, the perception layer processes 10 frames per second using YOLO-World, while visual anchors are selected according to the strategy in Sec.~\ref{sec:memory}. In practice, this results in one visual anchor being selected every 1-2 seconds to update the CMG, preventing blockage across system layers.
These results demonstrate that HiCo-Nav achieves real-time performance on embedded platforms, highlighting its practical efficiency and effectiveness in real-world robotic deployment.

\subsection{Ablation Study} \label{sec:ablation}

To evaluate the contribution of each component, we conduct an ablation study on the TextNav dataset, with results summarized in Table~V. The results show that removing the structural-semantic score $S_{\text{struct}}$ degrades performance (SR: 26.7\%, SPL: 12.1\%), highlighting the importance of modeling spatial-semantic correlations for exploration. Removing the out-of-boundary visual evidence $S_{\text{vis}}$ further reduces performance (SR: 26.5\%, SPL: 12.0\%), indicating the complementary role of long-range semantic cues beyond the explored region. Replacing the WTRP-based global planning with a greedy strategy leads to the largest drop (SR: 25.4\%, SPL: 11.8\%), demonstrating the necessity of global optimization for efficient long-horizon exploration.
Overall, the full system achieves the best performance (SR: 27.8\%, SPL: 12.9\%), confirming the effectiveness of the proposed methods.

\section{Conclusion}\label{sec:conclusion}

\begin{table}[t]
\centering
\caption{Ablation study.}
\begin{tabular}
{C{0.4\linewidth}|C{0.1\linewidth}C{0.1\linewidth}}
\toprule
\textbf{Method} & \textbf{SR} $\uparrow$ & \textbf{SPL} $\uparrow$ \\
\midrule
\textbf{Ours (full system)}                  & \textbf{27.8} & \textbf{12.9} \\
 w/o WTRP formulation                      & 25.4 & 11.8  \\
 w/o out-of-boundary score                 & 26.5 & 12.0 \\
 w/o structural-semantic score             & 26.7 & 12.1 \\
\bottomrule
\end{tabular}
\label{tab:ablation}
\end{table}


In this paper, we present HiCo-Nav, a deployable embodied vision-language navigation system that bridges the gap between high-level reasoning and real-time execution on resource-constrained robotic platforms. The proposed framework adopts a hierarchical and asynchronous design, decoupling perception, memory, and reasoning to enable both high-frequency responsiveness and expressive decision-making.
We introduce a cognitive memory graph that provides a compact and structured representation of the environment, supporting efficient long-horizon reasoning. Based on this representation, we develop a context-aware frontier exploration strategy that integrates structural-semantic propagation and out-of-boundary visual evidence to estimate exploration utility. Furthermore, we formulate goal selection as a Weighted Traveling Repairman Problem, enabling globally consistent exploration that minimizes expected time-to-discovery.
Extensive experiments in simulation and real-world environments demonstrate that HiCo-Nav achieves superior navigation performance and efficiency compared to existing methods, while maintaining real-time operation on embedded platforms.

\ifCLASSOPTIONcaptionsoff
  \newpage
\fi


{
    \bibliographystyle{IEEEtran}
    \bibliography{papers}
}


\vfill


\end{document}